\documentclass[10pt,twocolumn,letterpaper]{article}

\usepackage{cvpr}
\usepackage{times}
\usepackage{epsfig}
\usepackage{graphicx}
\usepackage{amsmath}
\usepackage{amssymb}

\usepackage{subcaption}
\usepackage{array}
\usepackage{multirow}
\usepackage{makecell}
\usepackage{tabularx}
\usepackage{booktabs}
\usepackage{tablefootnote}
\usepackage{caption}

\usepackage[pagebackref=true,breaklinks=true,letterpaper=true,colorlinks,bookmarks=false]{hyperref}

\cvprfinalcopy 

\def\cvprPaperID{5283} 

\ifcvprfinal\fi
\begin{document}
	
\title{MNEW: Multi-domain Neighborhood Embedding and Weighting for Sparse Point Clouds Segmentation}

\author{Yang Zheng, Izzat H. Izzat, Sanling Song\\
	Aptiv Corporation \\
	{\tt\small \{yang.zheng2, izzat.izzat, sanling.song\}@aptiv.com}
}

\maketitle

\begin{abstract}
	Point clouds have been widely adopted in 3D semantic scene understanding. However, point clouds for typical tasks such as 3D shape segmentation or indoor scenario parsing are much denser than outdoor LiDAR sweeps for the application of autonomous driving perception. Due to the spatial property disparity, many successful methods designed for dense point clouds behave depreciated effectiveness on the sparse data. In this paper, we focus on the semantic segmentation task of sparse outdoor point clouds. We propose a new method called MNEW, including multi-domain neighborhood embedding, and attention weighting based on their geometry distance, feature similarity, and neighborhood sparsity. The network architecture inherits PointNet which directly process point clouds to capture pointwise details and global semantics, and is improved by involving multi-scale local neighborhoods in static geometry domain and dynamic feature space. The distance/similarity attention and sparsity-adapted weighting mechanism of MNEW enable its capability for a wide range of data sparsity distribution. With experiments conducted on virtual and real KITTI semantic datasets, MNEW achieves the top performance for sparse point clouds, which is important to the application of LiDAR-based automated driving perception.
\end{abstract}


\begin{figure}[t]
	\centering
	\begin{center}
		\includegraphics[width=1.0\linewidth]{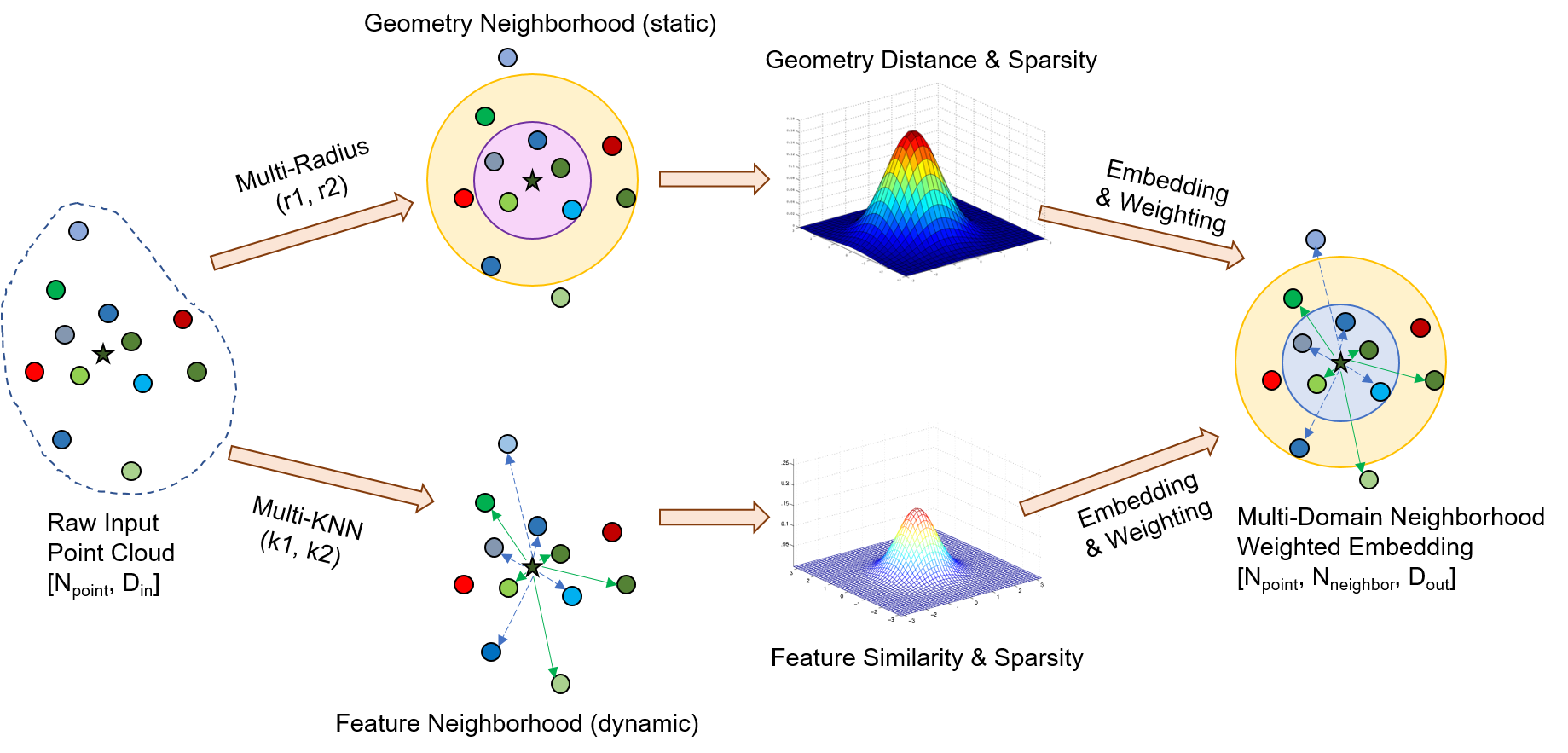}
	\end{center}
	\vspace{-0.2cm}
	\caption{Key idea illustration of MNEW. For a query point (star) within a set of point clouds (dots), we collect its static geometry neighbors by distance in two radius (upper branch), and its dynamic feature neighbors by similarity (shown in color) using two kNN (lower branch). Sparsities of each neighbor point are also computed in geometry and feature domain, which are used as weighting factors at the combined neighborhood embedding.}
	\label{fig1}
\end{figure}

\begin{figure}[t]
	\centering
	\begin{subfigure}[t]{0.45\textwidth}
		\centering
		\includegraphics[width=\textwidth, height=3.8cm]{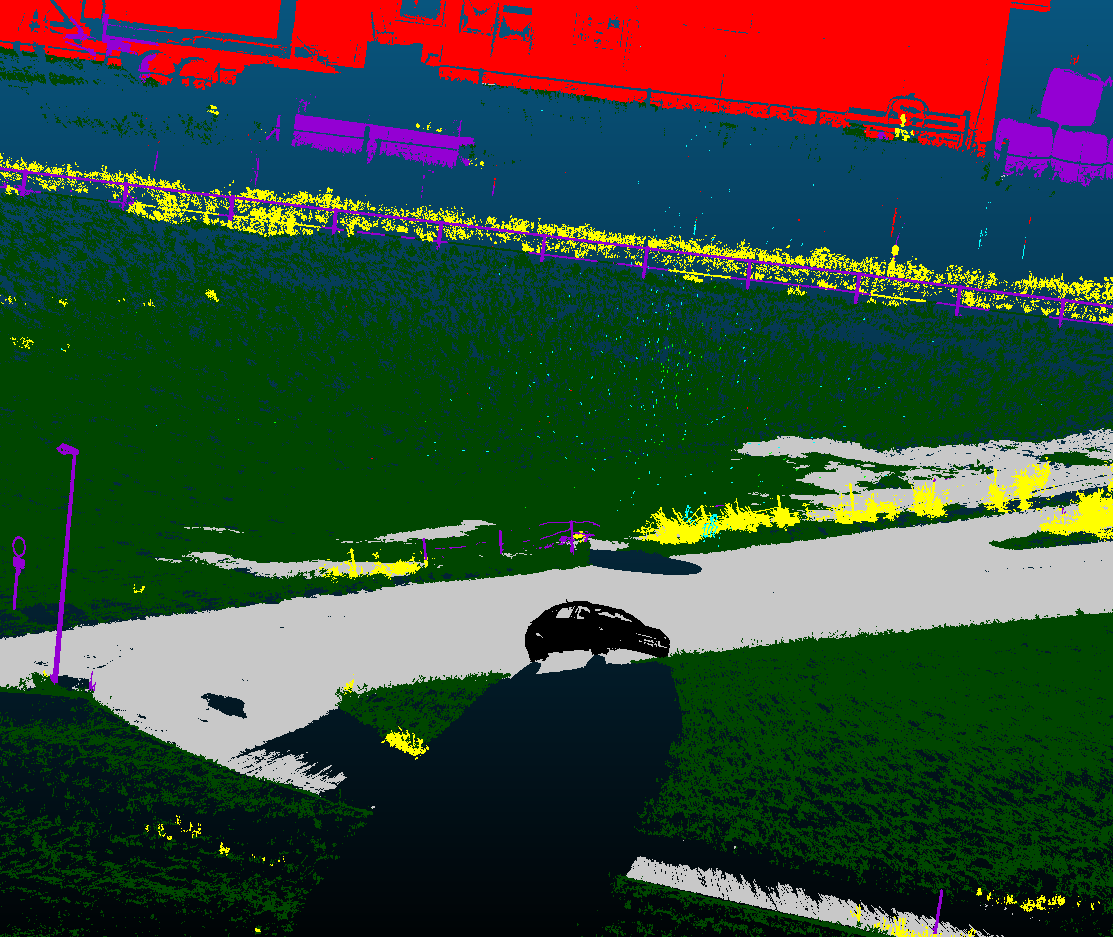}
		\vspace{-5mm}
		\caption{Dense Point Clouds}
		\label{fig2a}
	\end{subfigure}
	\begin{subfigure}[t]{0.45\textwidth}
		\centering
		\includegraphics[width=\textwidth, height=3.8cm]{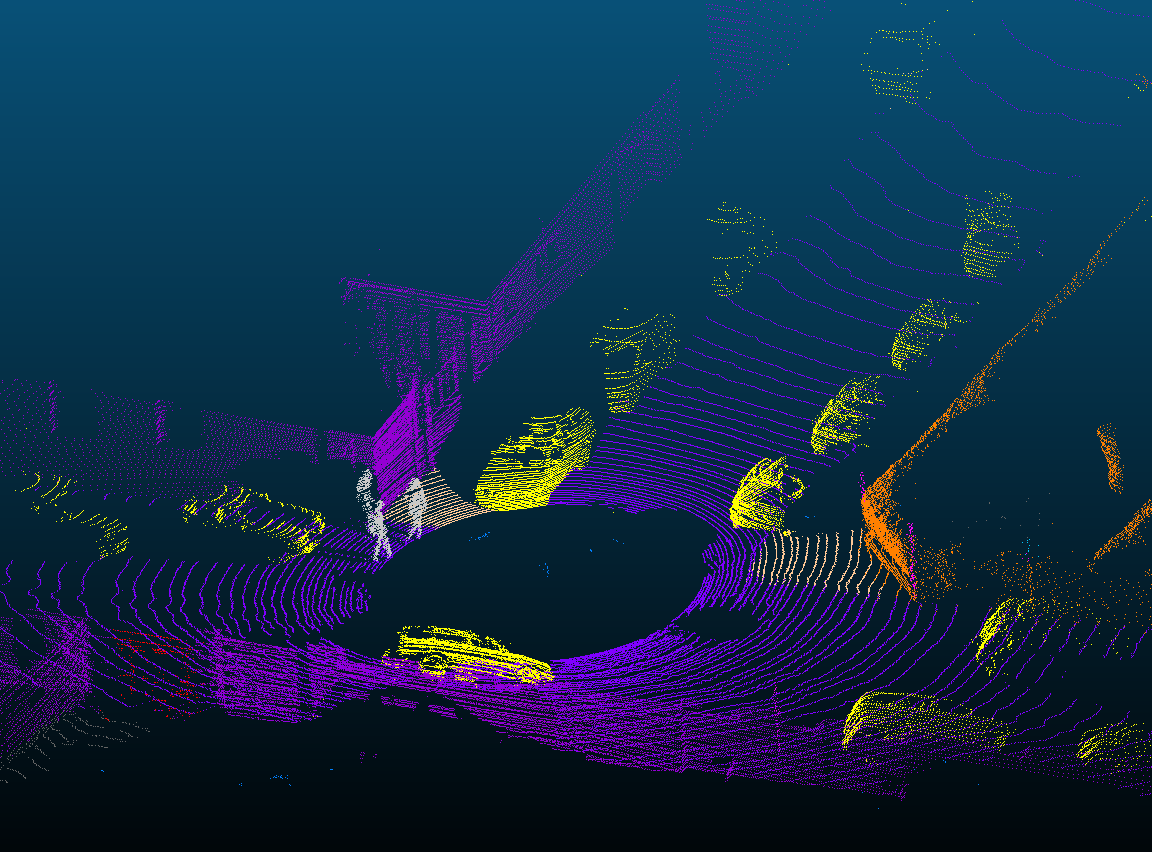}
		\vspace{-5mm}
		\caption{Sparse Point Clouds}
		\label{fig2b}
	\end{subfigure}
	\caption{Examples of (a) dense point clouds from Semantic3D \cite{hackel2017semantic3d}, and (b) sparse point clouds from KITTI \cite{geiger2012we}.}
	\label{fig2}
\end{figure}

\section{Introduction} \label{Introduction}

LiDAR point clouds, compared against other sensors such as camera and radar in the autonomous driving perception, have advantages of both accurate distance measurements and fine semantic descriptions. Studies on point clouds have gained increasing popularity in the computer vision area. Typical research topics include 3D shape recognition, part segmentation, indoor scenario parsing, and outdoor large-scale scene understanding. Several benchmark datasets such as ModelNet40 \cite{wu20153d}, ShapeNet \cite{chang2015shapenet}, S3DIS \cite{armeni20163d}, and Semantic3D \cite{hackel2017semantic3d} have been established for these topics. However, there exists spatial property disparity between these datasets and outdoor 360$^\circ$ sweep scans, which are typically produced by the on-board vehicle LiDAR. Per swept LiDAR point clouds are much sparser, and their sparsity is generally increased with the reflection distance. Examples in Figure \ref{fig2} demonstrate the difference between dense and sparse point clouds in outdoor areas.

Point clouds processing approaches could be generally classified into three categories. First, projection into 2D representations \cite{lawin2017deep, wu2018squeezeseg, caltagirone2017fast, meyer2019lasernet}. The basic idea of this approach is to transform unstructured point clouds into 2D images, which could be directly plugged into image-based Convolutional Neural Networks (CNN) \cite{krizhevsky2012imagenet}. Consequently, the 2D projection benefits to an easier fusion strategy with the image \cite{chen2017multi, meyer2019sensor} or multi-views \cite{su2015multi, qi2016volumetric}. However, the drawback is inevitably occluding a large number of points from the projection perspective, which suffers massive information loss. Second, voxelization into 3D volumetric grid cells \cite{maturana2015voxnet, huang2016point, li20173d, tchapmi2017segcloud, jiang2018pointsift} or their structural variations (e.g., Octree \cite{riegler2017octnet} or Spherical \cite{rao2019spherical}). Even though this approach is more effective to maintain points in the 3D space, the data loss problem is not eliminated due to the grid quantization. 3D convolutions are also computationally expensive. Third, directly process raw point clouds. A fundamental model PointNet \cite{qi2017pointnet} has been specifically designed to process raw unstructured point clouds, which inspires many other studies \cite{qi2017pointnet++, wang2018dynamic, wu2019pointconv, wang2019graph, wang2019associatively} to follow this idea. Many existing methods have reported their performance on several dense data classification and segmentation benchmarks, however, their effectiveness for sparse data is unknown.

Transfer learning \cite{pan2009survey} and domain adaptation \cite{tzeng2017adversarial} techniques have been recently discussed to bridge the gap between the source data and target data. They either extend the training procedure with new labeled data \cite{rist2019cross}, or re-design the network architecture by adopting an adversarial generator-discriminator \cite{lai2010object, wu2019squeezesegv2} for semi-supervised or unsupervised learning. In this particular study, however, we focus on the specific problem of semantic segmentation for sparse point clouds. We retain the cross-domain adaptation in our continued work. 

To the best of our knowledge and to the time of our work, VirtualKITTI \cite{3dsemseg_ICCVW17}, SemanticKITTI \cite{behley2019dataset}, and DeepenaiKITTI\footnote{https://www.deepen.ai/kitti-labels-download/} are currently available public datasets that provide semantic labels for sparse point clouds. We investigate a decent comparison of more than 10 state-of-the-art methods that directly process raw point clouds. With the evaluation of their effectiveness on the sparse data, we reveal that network architecture, neighborhood selection, and local sparsity are essential factors that affect the segmentation performance. This investigation has shown us advantages/shortcomings of previous methods, and therefore we are motivated to propose our method, Multi-domain Neighborhood Embedding and Weighting (MNEW). The key idea is illustrated in Figure \ref{fig1}. In MNEW, we collect multi-scale neighborhood points in both the static geometry domain and dynamic feature domain. Given a query point, its geometry neighbors are based on the Euclidean distance, and its feature neighbors are based on the similarity that are dynamically varied across different network layers. For each neighborhood point, we first assign attentions according to their location distance and feature similarity. We also compute the geometry/feature sparsity at each neighbor point, which are then transformed as adaptive weighting factors. The embedded neighborhood feature is a combination of weighted convolution outputs in the two domains. The overall network structure inherits PointNet, which is able to capture both pointwise details and global semantics. In addition, MNEW also extends its capability to embody the local contextual information. Experiments in Section \ref{Experiments} manifest the effectiveness of our method for the sparse data.

The major contributions of this paper are summarized as follows:
\begin{itemize}
	\itemsep 0em
	\item We introduce MNEW, a novel semantic segmentation model that is effective for per sweep LiDAR data, which is crucial for the application of autonomous driving perception.
	\item We design a neighborhood embedding method in both static geometry domain and dynamic feature space, which embodies attention mechanism by location distance and feature similarity, as well as sparsity-adapted weighting mechanism.
	\item We investigate a thorough comparison for a number of recent methods, evaluating their effectiveness on sparse point clouds. We claim that network architecture, neighborhood selection, and weighting mechanism are essential factors.
	\item We achieve state-of-the-art performance on sparse point clouds, and we observe that performance is varied by distance and local sparsity.
\end{itemize}


\begin{table*}[!htb]
	\begin{center}
		\footnotesize
		\begin{tabular}{c c c *3c c c}
			\toprule
			\multirow{2}{*}{Method}				& \multirow{2}{*}{Architecture} 	& \multirow{2}{*}{\makecell{Feature \\Extractor}}	& \multicolumn{3}{c}{Neighborhood}					& \multirow{2}{*}{Weighting}	& \multirow{2}{*}{Loss}							\\ \cline{4-6}
			&									&													& Domain			& Selection		& Embedding 	&								&												\\ 
			\midrule
			PointNet \cite{qi2017pointnet} 		& Dilated 							& MLP												& -					& - 			& Points		& -								& $\mathcal{L}_{CE}$							\\
			PointNet++ \cite{qi2017pointnet++}	& Encoder-Decoder 					& MLP												& Geometry			& Multi-radius	& Points 		& -								& $\mathcal{L}_{CE}$							\\
			A-CNN \cite{komarichev2019cnn}		& Encoder-Decoder 					& MLP												& Geometry			& Ring-shaped 	& Points		& -								& $\mathcal{L}_{CE}$							\\ 
			KP-FCNN \cite{thomas2019kpconv}		& Encoder-Decoder					& KP-Conv											& Geometry			& kNN in Radius	& Points		& Geometry Distance				& $\mathcal{L}_{CE} + \mathcal{L}_{Reg}$ 		\\
			DGCNN \cite{wang2018dynamic}		& Dilated							& MLP												& Feature			& kNN			& Query-Edges	& -								& $\mathcal{L}_{CE}$ 							\\
			RS-CNN \cite{liu2019relation}		& Encoder-Decoder					& RS-Conv											& Geometry			& Random-pick	& Query-Edges	& -								& $\mathcal{L}_{CE}$ 							\\
			PointWeb \cite{zhao2019pointweb}	& Encoder-Decoder					& MLP												& Geometry			& kNN			& Pairwise-Edges& -								& $\mathcal{L}_{CE}$ 							\\
			GACNet \cite{wang2019graph}			& Encoder-Decoder					& MLP												& Geometry			& Radius		& Points		& Feature Similarity			& $\mathcal{L}_{CE}$ 							\\
			PointConv \cite{wu2019pointconv}	& Encoder-Decoder					& MLP												& Geometry			& Radius		& Points		& Local Density					& $\mathcal{L}_{CE}$ 							\\
			ASIS \cite{wang2019associatively}	& Encoder-Decoder					& MLP												& Geometry			& Radius		& Points		& -								& $\mathcal{L}_{CE} + \mathcal{L}_{Disc}$		\\ 
			\hline
			\multirow{3}{*}{Ours (MNEW)}	& \multirow{3}{*}{\makecell{Dilated \\ (improved)}}	& \multirow{3}{*}{MLP}	& \multirow{3}{*}{\makecell{Geometry \\+ Feature}}	& \multirow{3}{*}{\makecell{Multi-radius \\+ Multi-kNN}}	& \multirow{3}{*}{Query-Edges}	& \multirow{3}{*}{\makecell{Geometry Distance \\+ Feature Similarity \\+ Neighbor Sparsity}}	& \multirow{3}{*}{$\mathcal{L}_{CE} + \mathcal{L}_{Reg}$} \\ 
			&									&													&					&				&				&								&												\\
			&									&													&					&				&				&								&												\\ 
			\bottomrule
		\end{tabular}
	\end{center}
	\vspace{-0.5cm}
	\caption{Comparison of methods that directly process raw point clouds.}
	\label{tab1}
\end{table*}

\section{Related Work} \label{Related Work}

\subsection{Methods} \label{sec2.1}
Since conversion-based approaches like 2D projection or 3D voxelization inevitably suffer the problem of losing points, in this section we focus on the related semantic segmentation methods that directly process raw point clouds, which are well-suited to explore the 3D data capability and close to our work.

PointNet \cite{qi2017pointnet} is considered a milestone method that inputs raw point clouds without any format transformation. This method adopts shared Multi-Layer Perceptrons (MLP) \cite{haykin1994neural} as the key component to learn pointwise features, and a pooling operation is followed to obtain global features representing all-points maximal response. The limit of PointNet is that it does not consider the local spatial relationship with neighborhood points. To address this issue, PointNet++ \cite{qi2017pointnet++} is proposed with a hierarchical encoder-decoder structure. Analogous to the Fully Convolutionally Networks (FCN) \cite{long2015fully} used in image segmentation, PointNet++ extracts local features by grouping and subsampling points in increasing contextual scales, and propagates subsampled features to their original points by interpolation.

Several subsequent studies improve PointNet and PointNet++ by their upgraded network designs. A-CNN \cite{komarichev2019cnn} introduces annular convolution with ring-shape neighborhoods to reduce the duplicated computation that exists in PointNet++ multi-scale grouping. Inspired from kernel pixels in the image-based convolution, KP-FCNN \cite{thomas2019kpconv} creates local 3D spatial filters using a set of kernel points. A function KP-Conv between kernel points and input points is defined, which is used to replace the MLP operation in PointNet/PointNet++. Alternative to the process of independent points, DGCNN \cite{wang2018dynamic} and RS-CNN \cite{liu2019relation} employ the idea of graph which embeds edges between a query point and its neighborhood points. The difference is that DGCNN follows the PointNet pipeline whereas RS-CNN follows the encoder-decoder structure of PointNet++. PointWeb \cite{zhao2019pointweb} extends this idea and proposes a pairwise edge embedding between every two points within the selected local region. Instead of the typical approach which collects neighborhood points based on their location distance, DGCNN collects neighbors in the dynamic feature space based on their similarity. Likewise, GACNet \cite{wang2019graph} assigns proper attention weights to different geometry neighbor points according to their feature attributes, which is to focus on the most relevant part of the neighbors. Weighting mechanism is also utilized in PointConv \cite{wu2019pointconv}, which estimates kernel density to re-weight the continuous function learned by MLP. SPG \cite{landrieu2018large} and its continued work \cite{landrieu2019point} partition point clouds into superpoint graphs and perform Edge-Conditioned Convolution (ECC) \cite{simonovsky2017dynamic} to assign a label on each superpoint. The difference lies in the graph construction approach, which is solved as an unsupervised minimal geometric partition problem in \cite{landrieu2018large} and a leaning-based method that minimizes the graph contrastive loss in \cite{landrieu2019point}. However, neither of these two methods is end-to-end. The purpose of graph contrastive loss is to detect the borders between adjacent objects. It pulls points belonging to the same object towards their centroid, while repelling those belonging to different objects. This idea is intuitively derived as the discriminative loss ($\mathcal{L}_{Disc}$) \cite{de2017semantic} in ASIS \cite{wang2019associatively}, which is added with the cross-entropy loss ($\mathcal{L}_{CE}$) and regularization loss ($\mathcal{L}_{Reg}$) for a joint semantic and instance segmentation.

Table \ref{tab1} summarizes an extensive comparison of selected methods, which are varied by the network architecture, feature extractor, neighborhood selection/embedding, weighting mechanism, and loss function. Our proposed method is also listed to overview the relation and distinction. More details are discussed in Section \ref{Method}.

\begin{table*}[!htp]
	\begin{center}
		\footnotesize
		\begin{tabular}{c c c c c c c c}
			\toprule
			Dataset									& Type 				& Attributes	& Size (Train + Test)					& Classes				& Instance	& Sequential	& Train/Valid 			\\
			\midrule
			S3DIS \cite{armeni20163d}				& indoor dense		& XYZ + RGB		& 6 indoor area, 273M points			& 13					& Yes		& No			& -						\\
			ScanNet \cite{dai2017scannet}			& indoor dense		& XYZ + RGB		& 1.5K scans, 2.5M frames				& 21					& Yes		& Yes			& -						\\
			Semantic3D \cite{hackel2017semantic3d}	& outdoor dense		& XYZ + RGB		& 30 scenarios, 4009M points			& 9						& No		& No			& -						\\
			NPM3D \cite{roynard2018paris}			& outdoor dense		& XYZ			& 6 scenarios, 143M points				& 50 (10)				& Yes		& No			& -						\\
			\midrule
			VirtualKITTI \cite{3dsemseg_ICCVW17}	& outdoor sparse	& XYZ + RGB		& 4 simulated scenes, 90 frames			& 14					& No		& No			& 80\%/20\%	random		\\
			DeepenaiKITTI							& outdoor sparse	& XYZ			& 1 sequence, 100 frames				& 17					& No		& Yes			& 80\%/20\% random		\\
			SemanticKITTI \cite{behley2019dataset}	& outdoor sparse	& XYZ			& 22 sequences, 43.5K frames			& 28 (20)				& Yes		& Yes		 	& 10/1 sequence			\\
			\bottomrule
		\end{tabular}
	\end{center}
	\vspace{-0.5cm}
	\caption[Caption for LOF]{Comparison of selected datasets with dense and sparse point clouds.}
	\label{tab2}
\end{table*}

\begin{table*}[!htb]
	\begin{center}
		\footnotesize
		\begin{tabular}{c| *2c *2c *2c| *2c *2c *2c}
			\toprule
			\multirow{2}{*}{Method}				& \multicolumn{2}{c}{S3DIS} 	& \multicolumn{2}{c}{ScanNet}	& \multicolumn{2}{c|}{Semantic3D}		& \multicolumn{2}{c}{VirtualKITTI}	& \multicolumn{2}{c}{DeepenaiKITTI}	& \multicolumn{2}{c}{SemanticKITTI}		\\ \cline{2-13}
												& OA			& mIoU			& OA			& mIoU			& OA			& mIoU					& OA			& mIoU				& OA			& mIoU				& OA			& mIoU					\\
			\midrule
			PointNet \cite{qi2017pointnet}		& 78.62			& 47.71			& 73.9			& -				& -				& -						& 88.07			& 50.36				& 98.41			& 64.85				& 66.12			& 19.74					\\
			PointNet++ \cite{qi2017pointnet++}	& -				& -				& 84.5			& 33.9			& 82.5			& 52.1					& 81.99			& 44.74				& 96.66			& 54.40				& 72.35			& 22.90					\\
			A-CNN \cite{komarichev2019cnn}		& 87.3			& 62.9			& \textbf{85.4}	& -				& -				& -						& 42.80			& 18.75				& 43.15			& 7.31				& 33.35			& 7.85					\\
			KP-FCNN \cite{thomas2019kpconv}		& -				&\textbf{65.4}	& -				& \textbf{68.6}	& \textbf{92.9}	& \textbf{74.6}			& 75.02			& 30.49				& 36.75			& 4.54				& 78.05			& 26.71					\\
			DGCNN \cite{wang2018dynamic}		& 84.1			& 56.1			& -				& -				& -				& -						& 92.04			& 60.19				& 98.28			& 64.54				&\textbf{80.64}	&\textbf{30.51}			\\
			PointWeb \cite{zhao2019pointweb}	& 86.97			& 60.28			& 85.9			& -				& -				& -						& 57.06			& 18.94				& 67.98			& 16.67				& 32.17			& 6.84					\\
			GACNet \cite{wang2019graph}			&\textbf{87.79}	& 62.85			& -				& -				& 91.9			& 70.8					&\textbf{92.57}	&\textbf{60.58}		& 95.56			& 51.38				& 76.51			& 26.06					\\
			PointConv \cite{wu2019pointconv}	& -				& -				& -				& 55.6			& -				& -						& 85.26			& 47.60				&\textbf{98.50}	&\textbf{65.74}		& 72.51			& 23.24					\\
			\bottomrule
		\end{tabular}
	\end{center}
	\vspace{-0.5cm}
	\caption{Comparison of existing methods on dense and sparse point clouds.}
	\label{tab3}
\end{table*}


\begin{figure*}[t]
	\centering
	\begin{center}
		\includegraphics[width=0.9\linewidth]{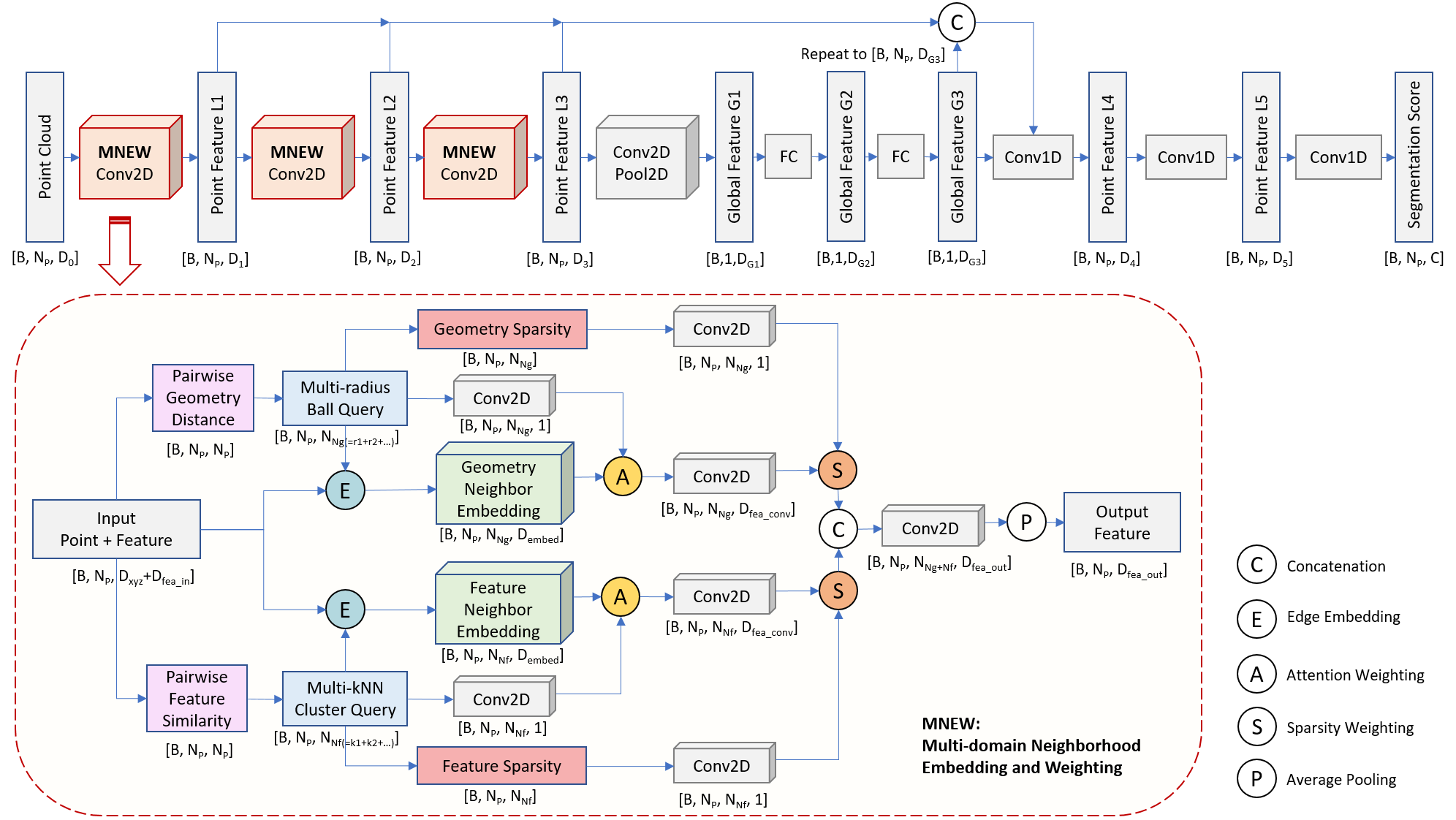}
	\end{center}
	\caption{Design of our proposed network. The key module MNEW is zoomed in for a detailed illustration. In MNEW, the upper branch embeds static geometry neighbors based on their location distance (by multi-radius), and the lower branch embeds dynamic feature neighbors based on their similarity (by multi-kNN). Local sparsity is computed in both geometry and feature domain, and transformed to weight the convolution output. After concatenation, a pooling operation aggregates neighborhood features for each query point.}
	\label{fig3}
\end{figure*}


\subsection{Datasets} \label{sec2.2}

For the task of 3D scene semantic segmentation, publicly available datasets such as S3DIS \cite{armeni20163d} and ScanNet \cite{dai2017scannet} are indoor dense data, whereas Semantic3D \cite{hackel2017semantic3d} and NPM3D \cite{roynard2018paris} are outdoor. For all of the four benchmark datasets, point clouds are collected by accumulating multiple scans in stationary environments to obtain fine detailed measurements. However, sparse point clouds for the application of autonomous driving perception, like the example shown in Figure \ref{fig2b}, are much different. A sweeping LiDAR sensor is mounted on a moving vehicle, correspondingly the scanning environment is also changing. In a single frame, point clouds are generally denser in close areas, and much sparser far away from the sensor. This is determined by hardware characteristics of the rotation LiDAR sensor. To the time of our work, we found three public datasets with sparse point clouds. VirtualKITTI \cite{3dsemseg_ICCVW17} is simulated virtual data, while DeepenaiKITTI and SemanticKITTI \cite{behley2019dataset} provide small/large-sized semantic labels on the real world KITTI \cite{geiger2012we} sequential data.

Table \ref{tab2} summarizes the selected datasets. Other popular benchmarks such as ModelNet40 \cite{wu20153d} or ShapeNet \cite{chang2015shapenet} are focused on small 3D CAD objects, which are beyond our interest scope. Note for the number of classes, NPM3D has 50-class fine annotations and 10-class coarse annotations; SemanticKITTI has 28-class labels to separate moving/stationary objects, and mapped into 20-class for closest equivalent. Since labels for their test subsets are invisible before submission, we split the annotated training data into training/validation in our experiments for verification.

\subsection{Analysis} \label{sec2.3}

Table \ref{tab3} compares the performance of selected methods on selected datasets. For dense point clouds, results are borrowed from their original reports. For sparse data, we use the train/valid splits in Table \ref{tab2}, and re-produce experiments with their official implementations. Evaluation metrics are the overall accuracy (OA) for every point and the mean intersection-over-union (mIoU) for averaged IoU across all classes. From Table \ref{tab3}, we summarize our findings as follows.

{\bf Architecture.} 
PointNet and DGCNN are the networks retaining the number of points unchanged, while other methods are hierarchical encoder-decoders where points are down-sampled and then up-sampled. Despite the fact that encoder-decoder networks performs better in dense data benchmarks, their effectiveness is depreciated for sparse point clouds. One possible explanation is that, 3D interpolations for up-sampling might be suitable for the near-uniformly distributed dense point clouds, but not for the irregular sparse data. Since no down-sample/up-sampling exists in PointNet and DGCNN, they are similar to the dilated convolution \cite{yu2015multi} in the image segmentation, which maintains resolution unchanged end-to-end for element-wise feature learning. 

{\bf Neighborhood.} 
DGCNN selects neighboring points in the dynamic feature domain, which differs from other methods whose neighbors are selected by the static geometry distance. GACNet collects geometry neighbors but assigns attention weights based on their feature similarity. Since the performances of DGCNN and GACNet seem promising for SemanticKITTI and VirtualKITTI, we infer that dynamic feature-based neighborhoods are critical. This is interpretable, for example, sparse points on the road in far distances are isolated in geometry location, but they are similar in their feature representations. In contrast, traffic signs hidden in trees are close to leaves, but they should be distinguished from the surrounding vegetation.

{\bf Weighting.} 
Similar to the attention schema in GACNet, PointConv compromises the density estimation as a weighting function, which is learned to adapt with different local densities. This method directly considers the data density and therefore obtaining encouraging performance on DeepenaiKITTI. We infer that weighting mechanisms in GACNet and PointConv could compensate for the effectiveness depreciation of their encoder-decoder architecture. 

{\bf Dataset Discrepancy.} 
According to the experimental results in Table \ref{tab3}, DGCNN, GACNet and PointConv are the preferred methods on sparse point clouds. However, the performance is inconsistent across the three selected sparse datasets. The major reason is essentially the data intrinsic discrepancy. VirtualKITTI is generated by the simulator, and it is the only sparse data with RGB colors. SemanticKITTI is a large-scale sequential dataset, which suggests 10 sequences for training and 1 other sequence for validation. DeepenaiKITTI is small-sized probe data, and all its currently available frames are extracted from the same sequence. Due to the small size and high correlation of DeepenaiKITTI, we neglect it in Section \ref{Experiments} and evaluate our method on VirtualKITTI and SemanticKITTI.


\begin{table*}[!htb]
	\scriptsize
	\centering
	\newcolumntype{M}{ >{\arraybackslash} p{0.27cm} }
	\begin{subtable}[!htb]{\textwidth}
		\centering
		\begin{tabular}{c| M M M M M M M M M M M M M M| M M}
			\toprule
			Method	&\rotatebox[origin=c]{90}{Terrian}	&\rotatebox[origin=c]{90}{Tree}	&\rotatebox[origin=c]{90}{Vegetation}	&\rotatebox[origin=c]{90}{Building}	&\rotatebox[origin=c]{90}{Road}	&\rotatebox[origin=c]{90}{Guardrail}	&\rotatebox[origin=c]{90}{Traffic Sign}	&\rotatebox[origin=c]{90}{Traffic Light}	&\rotatebox[origin=c]{90}{Pole}	&\rotatebox[origin=c]{90}{Misc}	&\rotatebox[origin=c]{90}{Truck}	&\rotatebox[origin=c]{90}{Car}	&\rotatebox[origin=c]{90}{Van}	&\rotatebox[origin=c]{90}{Unlabeled} 	& OA	& mIoU	\\
			\midrule
			DGCNN \cite{wang2018dynamic}		& 86.7			& 92.5			& 70.8			& 81.2			& 94.8			& 93.9			& 38.0			& 78.0			& 65.5			& 27.3			& 29.2			& 76.3			& 8.6			& 0.0	& 92.0			& 60.2	\\
			GACNet \cite{wang2019graph}			& 82.0			& 95.2			& 72.5			& 86.6			& 92.1			& 90.6			& 51.4			& 48.1			& 42.6			& 31.1			& 46.6			& 81.6			& \textbf{27.8}	& 0.0	& 92.6			& 60.6	\\
			PointConv \cite{wu2019pointconv}	& 58.1			& 89.4			& 57.0			& 76.0			& 80.6			& 66.9			& 25.2			& 59.3			& 25.1			& 35.6			& 9.14			& 72.4			& 12.0			& 0.0	& 85.3			& 47.6	\\
			\midrule
			MNEW-4096							& 92.6			& \textbf{97.7}	& 84.5			& 90.7			& 97.6			& 97.3 			& 68.8			& 71.9			& 62.6			& 52.9			& 11.0			& 85.9			& 23.5			& 0.0	& 95.9			& 67.0	\\
			MNEW-2048							& \textbf{97.0}	& \textbf{97.7} & \textbf{91.2} & \textbf{92.4} & \textbf{98.8} & \textbf{98.2} & \textbf{70.6} & \textbf{83.8} & \textbf{72.8} & \textbf{64.9} & \textbf{58.4} & \textbf{88.3} & 12.7			& 0.0	& \textbf{97.1} & \textbf{73.3}	\\
			\bottomrule
		\end{tabular}
		\vspace{-0.1cm}
		\caption{Validation results on VirtualKITTI dataset}
		\label{tab4a}
		\vspace{0.1cm}
	\end{subtable}

	\begin{subtable}[!htb]{\textwidth}
		\centering
		\begin{tabular}{c| M M M M M M M M M M M M M M M M M M M M| M M}
			\toprule
			Method	&\rotatebox[origin=c]{90}{Car}	&\rotatebox[origin=c]{90}{Bicycle}	&\rotatebox[origin=c]{90}{Motorcyclist}	&\rotatebox[origin=c]{90}{Truck}	&\rotatebox[origin=c]{90}{Other Vehicle}	&\rotatebox[origin=c]{90}{Person}	&\rotatebox[origin=c]{90}{Bicyclist}	&\rotatebox[origin=c]{90}{Motorcyclist}	&\rotatebox[origin=c]{90}{Road}	&\rotatebox[origin=c]{90}{Parking}	&\rotatebox[origin=c]{90}{Sidewalk}	&\rotatebox[origin=c]{90}{Other Ground}	&\rotatebox[origin=c]{90}{Building}	&\rotatebox[origin=c]{90}{Fence}	&\rotatebox[origin=c]{90}{Vegetation}	&\rotatebox[origin=c]{90}{Trunk}	&\rotatebox[origin=c]{90}{Terrain}	&\rotatebox[origin=c]{90}{Pole}	&\rotatebox[origin=c]{90}{Traffic Sign}	&\rotatebox[origin=c]{90}{Unlabeled}	& OA	& mIoU \\
			\midrule
			DGCNN \cite{wang2018dynamic}		& 78.3			& 0.0			& 1.1			& \textbf{17.3}	& 1.4			& 1.9			& 3.9			& 0.0	& 88.7			& 10.2			& 65.1			& 0.1			& 74.1			& 18.8			& 71.6			& 25.0			& 62.1			& 28.5			& 8.8			& 47.8			& 80.8			& 30.2			\\
			GACNet \cite{wang2019graph}			& 71.5			& 0.0			& 0.0 			& 12.2			& 1.4			& 0.0			& 0.0			& 0.0	& 80.3			& 13.4			& 55.3			& 0.2			& 63.1			& 16.7			& 67.8			& 15.7			& 56.4			& 12.1			& \textbf{22.9}	& 38.8			& 76.0 			& 26.4 			\\
			PointConv \cite{wu2019pointconv}	& 60.5			& 0.1			& 0.2			& 0.6			& 3.3			& 1.0			& 0.9			& 0.0	& 82.1			& 3.8			& 55.4			& \textbf{0.4}	& 63.6			& 10.8			& 59.6			& 14.2 			& 52.1			& 14.1			& 8.0			& 34.4			& 72.5			& 23.2			\\
			RangeNet \cite{milioto2019rangenet++}	& 74.1		& \textbf{14.3}	& 2.6			& 9.4			& 10.5			& \textbf{7.2}	& 21.9			& 0.0	& \textbf{90.7}	& \textbf{36.2}	& \textbf{74.2}	& 0.2			& 67.8			& \textbf{33.4}	& 71.9			& 30.7			& \textbf{68.5}	& 23.0			& 22.2			& 34.1			& 81.4			& 34.6 			\\
			\midrule
			MNEW-4096							& \textbf{81.3}	& 0.0			& \textbf{13.3}	& 8.8			& \textbf{12.9}	& 6.2			& \textbf{31.7}	& 0.0	& 88.7			& 22.3			& 70.4			& 0.1			& \textbf{79.3}	& 30.0			& \textbf{76.9}	& \textbf{34.2}	& 66.4			& \textbf{33.1}	& 1.1			& \textbf{49.3}	& \textbf{84.1}	& \textbf{35.3} \\
			MNEW-2048							& 79.8			& 0.0			& 10.5			& 6.5			& 7.8			& 5.5			& 25.5			& 0.0	& 88.8			& 22.7			& 67.4			& 0.0			& 77.2			& 29.1			& 75.0			& 29.6			& 61.9			& 27.3			& 1.4 			& 47.5			& 82.5			& 33.2 			\\
			\bottomrule
		\end{tabular}
		\vspace{-0.1cm}
		\caption{Validation results on SemanticKITTI dataset}
		\label{tab4c}
		\vspace{0.1cm}
	\end{subtable}
	
	\caption{Semantic segmentation results on sparse point clouds. Metrics are OA(\%), mIoU(\%), and per class IoU(\%).}
	\label{tab4}
	\end{table*}


\section{Methodology} \label{Method}

\subsection{Network Architecture} \label{sec3.1}

Inspired from the findings in Section \ref{sec2.3}, we propose our network design illustrated in Figure \ref{fig3}. The overall architecture inherits a dilated structure like PointNet \cite{qi2017pointnet} or DGCNN \cite{wang2018dynamic}, which eliminates re-sampling operations end-to-end. The model takes batches of input $N_p$ points, and passes through a sequence of three MNEW modules to extract pointwise features $L_1$, $L_2$, and $L_3$ in a hierarchical order. Note that local neighborhood information is carried inside MNEW, which yields the upgrade from PointNet. We increase the number of neighbors in $L_1$, $L_2$, and $L_3$, which correspond to hierarchical-scale feature encoders but keep the number of query points fixed. A global 2D convolution, max pooling, and two fully convolution layers are followed to aggregate the global feature $G_3$. The hierarchical pointwise features and tiled global feature are concatenated as a descriptor for each point, and passed through three regular 1D convolutions to get the segmentation score, i.e., category-level probability. 

\subsection{Multi-domain Neighborhood Embedding and Weighting} \label{sec3.2}

The key component in our network design is the multi-domain neighborhood embedding and weighting (MNEW) module. As shown in Figure \ref{fig3}, the input of MNEW is batches of points with their xyz coordinates and original features, shaped as $[B, N_p, D_{xyz+fea}]$. We compute pairwise distances in both geometry and feature domain, resulting in two matrices with shape $[B, N_p, N_p]$ representing the geometry distance and feature similarity between every two points. 

As discussed in KP-Conv \cite{thomas2019kpconv} and RS-Conv \cite{liu2019relation}, radius-based geometry neighborhood selection is more robust than k-NN \cite{weinberger2006distance} in the non-uniform sampling settings. However, feature neighborhoods are dynamically shifting and hard to be encircled. Therefore, we use multiple radius in the geometry domain and multiple k-NN in the feature domain to collection multi-scale neighbor indices. We gather their original features to compose two initial neighborhood embedding matrices $\mathbf{X}_g^0$ and $\mathbf{X}_f^0$ with shape $[B, N_p, N_{ng}, D_{embed}]$ and $[B, N_p, N_{nf}, D_{embed}]$ respectively, where $ng=\sum r_i$ represents the number of accumulated neighbors in multi-radius geometry space, and $nf=\sum k_i$ represents the number of accumulated neighbors in multi-kNN feature space. Similar to the graph embedding $G(V,E)$ in DGCNN and RS-Conv which includes vertices and edges, we embed each element in $\mathbf{X}_g^0$ and $\mathbf{X}_f^0$ as,
\begin{equation} \label{eq1}
\mathbf{f}(x_{i,n_j}) = \mathbf{f}(x_{n_j}, x_{n_j}-x_i), \quad x_{i,n_j} \in (\mathbf{X}_g^0 \cup \mathbf{X}_f^0)
\end{equation}
where $x_i$ denotes the $i$-th query point, and $x_{n_j}$ denotes the $j$-th neighbor point. Given indices of selected neighbors, we also gather their geometry distance matrix $\mathbf{D}_g$ (shape $[B, N_p, N_{ng}, 1]$) and feature similarity matrix $\mathbf{D}_f$ (shape $[B, N_p, N_{nf}, 1]$). Next, a transformation function $\mathbf{T}(d_{i,n_j})=\mathbf{w}_{i,n_j}\cdot\mathbf{f}(d_{i,n_j})$ is utilized to obtain adaptive attention weights. The attended embedding $\mathbf{X}_g^a$ and $\mathbf{X}_f^a$ are computed as, 
\begin{equation} \label{eq2}
\begin{split}
\mathbf{X}_g^a = \mathbf{T}(\mathbf{D}_g) \cdot \mathbf{X}_g^0 \\
\mathbf{X}_f^a = \mathbf{T}(\mathbf{D}_f) \cdot \mathbf{X}_f^0
\end{split}
\end{equation}

Motivated by the density estimation in PointConv \cite{wu2019pointconv}, we calculate the neighborhood sparsity using,
\begin{equation} \label{eq3}
\mathbb{P}(x_{i,n_j}|\mu, \sigma^2) = \frac{1}{\sqrt{2\pi\sigma^2}} \exp{[-\frac{(x_{n_j}-x_i)^2}{2\sigma^2}]}
\end{equation}
\begin{equation} \label{eq4}
\mathbb{S}(x_i|\mu, \sigma^2) = (\frac{1}{N_n}\log[\sum_{n_j \in N_n} \mathbb{P}(x_{i,n_j}|\mu, \sigma^2)])^{-1}
\end{equation}
$\mathbb{P}(x_{i,n_j}|\mu, \sigma^2)$ in Equation (\ref{eq3}) is equivalent to the Gaussian probability density function computed for every neighbor $x_{n_j}$ with respect to the query point $x_{i}$. $\mathbb{S}(x_{i,n_j})$ in Equation (\ref{eq4}) is the estimated sparsity which inverses the averaged density. We also take the log-scale value to obtain a better sparsity distribution (see Figure \ref{fig4b}). Geometry sparsity $\mathbb{S}_g$ and feature sparsity $\mathbb{S}_f$ are computed individually, which are transformed as the weighting factor for the 2D convolution activation $\mathbf{h}(x)$. The weighted outputs $\mathbf{X}_g^w$ (shape $[B, N_p, N_{ng}, D_{conv}]$) and $\mathbf{X}_f^w$ (shape $[B, N_p, N_{nf}, D_{conv}]$) are computed as,
\begin{equation} \label{eq5}
\begin{split}
\mathbf{X}_g^w = \mathbf{T}(\mathbb{S}_g) \cdot \mathbf{h}(\mathbf{X}_g^a) \\
\mathbf{X}_f^w = \mathbf{T}(\mathbb{S}_f) \cdot \mathbf{h}(\mathbf{X}_f^a)
\end{split}
\end{equation}
After concatenating the neighborhood information from geometry and feature domain, an average pooling operation is followed to aggregate a feature vector for each query point, yielding the output of MNEW module $\mathbf{X}_{mnew}^{out}$ with shape $[B, N_p, D_{conv}]$.
\begin{equation} \label{eq6}
\mathbf{X}_{mnew}^{out} = \frac{1}{N_p} \sum_{i \in N_p} (\mathbf{X}_g^w \oplus \mathbf{X}_f^w)
\end{equation}

\subsection{Loss Function} \label{sec3.3}

The loss function is a combination of softmax cross-entropy loss $\mathcal{L}_{CE}$ and regularization loss $\mathcal{L}_{Reg}$ adjusted by $\lambda$. Since the task is semantic segmentation only (i.e., no instance-level labels), the discriminitive loss suggested by ASIS \cite{wang2019associatively} is not applicable.
\begin{equation} \label{eq7}
\mathcal{L}_{Total} = \mathcal{L}_{CE} + \lambda \mathcal{L}_{Reg}
\end{equation}


\subsection{Comparison to Existing Methods} \label{sec3.4}

Referring to Table \ref{tab1}, we compare existing methods and summarize the essential distinctions of MNEW as follows.

The dilated network architecture in our work excludes downsample grouping and upsample interpolation, which differs from all recent works that are based on hierarchical encoder-decoder structures. Compared with PointNet \cite{qi2017pointnet} whose feature contains pointwise and global information, we also include local neighborhood features. Compared with DGCNN \cite{wang2018dynamic} which collects neighbors in the feature space only, we embed neighbor points in both geometry and feature domain.

In terms of the neighborhood embedding, our method adopts multi-scaling in multi-domain. This differs from all existing methods where neighbors are collected in only one single domain (i.e., either geometry or feature). Hierarchical PointNet++ \cite{qi2017pointnet++} and A-CNN \cite{komarichev2019cnn} use multi-radius ball-shaped or ring-shaped scales in the geometry domain, while DGCNN using single-scale kNN in the feature domain. In our method, there may exist overlapping points selected in geometry and feature neighborhoods. However, since we compute adaptive attention \& weighting factors in each domain separately, their impact are learned individually.

For the attention/weighting mechanism, KP-FCNN \cite{thomas2019kpconv} and GACNet \cite{wang2019graph} compute the geometry distance or feature similarity as fixed weighting factors, while PointConv \cite{wu2019pointconv} transforms the local sparsity as a learning-based flexible variable. In our method, all these factors are trainable.


\section{Experiments} \label{Experiments}

\subsection{Sparse Point Cloud Segmentation} \label{sec4.1}

Experimental results on VirtualKITTI and SemanticKITTI are shown in Table \ref{tab4}, using the train/valid splits from Table \ref{tab2}. Evaluation metrics include OA, mIoU, and per class IoU. We select DGCNN, GACNet, and PointConv as baselines since their accuracies are higher than other related methods (see Table \ref{tab3}). Our proposed method, MNEW, achieves outstanding performances on both VirtualKITTI and SemanticKITTI. We experimentally set the number of points (i.e., $N_p$) 4096 or 2048, since it affects the global feature extraction and neighborhood selection. Table \ref{tab4} indicates that MNEW-2048 performs better on VirtualKITTI, while MNEW-4096 is superior on SemanticKITTI. Comparing against the top-performed baseline method, MNEW facilitates 4.5\% OA and 12.7\% mIoU increments on VirtualKITTI, a greater margin than those on SemanticKITTI (3.3\% OA and 5.1\% mIoU higher than DGCNN). This is because VirtualKITTI provides RGB information as its original feature, which is independent from the XYZ geometry coordinates. For SemanticKITTI, we use intensity, and compute 2D\_range and 3D\_distance to fill the input feature slots. Therefore, the multi-domain neighborhood combination is more effective for VirtualKITTI. 

SemanticKITTI contains more categorical labels, but the percentage of unlabeled points (class-0) is also higher than VirtualKITTI (4.49\% vs. $<$0.01\%). For SemanticKITTI, our experimental results slightly differ from those reported in \cite{behley2019dataset}. One of the critical reason is that unlabeled points, despite occupying considerable large percentage, are ignored in \cite{behley2019dataset}. With the class-0 included, we reproduced the RangeNet \cite{milioto2019rangenet++}, an upgraded DarkNet in \cite{behley2019dataset}. We provide a consistent comparison as listed in Table \ref{tab4c}. In addition to DarkNet/RangeNet, \cite{behley2019dataset} and \cite{milioto2019rangenet++} also claim that projection-based methods such as SqueezeSeg \cite{wu2018squeezeseg} or TangentConv \cite{tatarchenko2018tangent} are more effective than those directly processing raw data. Since per sweep point clouds are collected by a rotational LiDAR, a cylinder-view projection exactly looks at those points from the perspective of LiDAR sensor. This could ensure its effectiveness for small objects, such as bicycles and traffic-signs which are better viewed from the sensor's perspective. Although reasonable, MNEW still obtains higher mIoU (35.3\% vs. 34.6\%) and OA (84.1\% vs. 81.4\%) than RangeNet.

\begin{table}[t]
	\begin{center}
		\scriptsize
		\begin{tabular}{c c c c| c c}
			\toprule
			Geometry		& Feature			& Distance/Similarity	& Sparsity		& OA		& mIoU		\\
			\midrule
			\checkmark		&					&						&				& 92.1		& 57.0		\\
							& \checkmark		&			 			&				& 95.8		& 65.9		\\
			\checkmark		& \checkmark		&			 			&				& 95.9		& 69.7		\\
			\midrule
			\checkmark		& \checkmark		& \checkmark 			&				& 96.3		& 72.6		\\
			\checkmark		& \checkmark		&  						& \checkmark	& 96.9		& 70.6		\\
			\checkmark		& \checkmark		& \checkmark 			& \checkmark	& 97.1		& 73.3		\\
			\bottomrule
		\end{tabular}
	\end{center}
	\vspace*{-1mm}
	\caption{Effect of neighborhood selection, distance/similarity attention, and sparsity weighting. Experimented MNEW-2048 on VirtualKITTI}
	\label{tab5}
\end{table}

\begin{figure*}[t!]
	\centering
	\begin{subfigure}[t]{\textwidth}
		\centering
		\begin{subfigure}[t]{0.4\textwidth}
			\includegraphics[width=\textwidth, height=3.8cm]{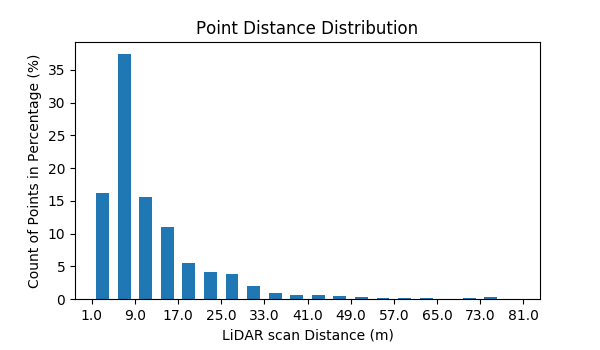}
		\end{subfigure}
		\begin{subfigure}[t]{0.5\textwidth}
			\includegraphics[width=\textwidth, height=3.8cm]{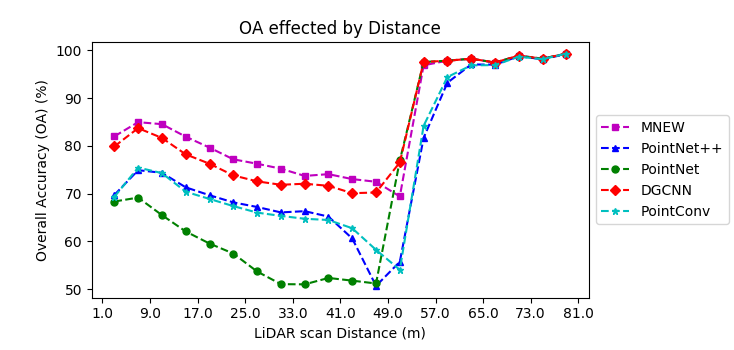}
		\end{subfigure}
		\vspace*{-1mm}
		\caption{Distance (m) Distribution and OA (\%) by Distance Effects}
		\label{fig4a}
	\end{subfigure}
	\begin{subfigure}[t]{\textwidth}
		\centering
		\begin{subfigure}[t]{0.4\textwidth}
			\includegraphics[width=\textwidth, height=3.8cm]{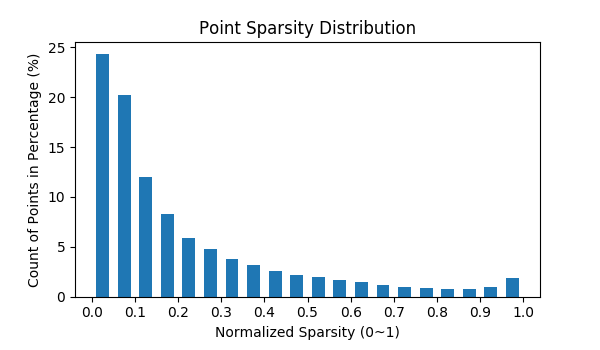}
		\end{subfigure}
		\begin{subfigure}[t]{0.5\textwidth}
			\includegraphics[width=\textwidth, height=3.8cm]{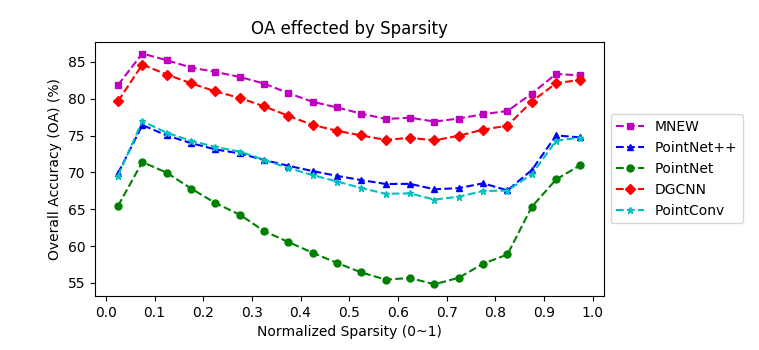}
		\end{subfigure}
		\vspace*{-1mm}
		\caption{Sparsity (normalized 0-1) Distribution and OA (\%) by Sparsity Effects}
		\label{fig4b}
	\end{subfigure}
	\caption{Points distribution and performance variation by distance and sparsity. Experimented MNEW-4096 on SemanticKITTI.}
	\label{fig4}
\end{figure*}

\subsection{Ablation Studies} \label{sec4.2}

{\bf Neighborhood Embedding and Weighting.}
In Table \ref{tab5}, we compare several variations of the model design. For the neighborhood selection, we compare the neighbors collected by geometry location, feature similarity, and both. The experiment reveals that the accuracy significantly increased with the supplement of feature domain neighborhood. Next, we optionally enable the neighbor attention of geometry distance and feature similarity ($\mathbf{T}(\mathbf{D}_g)$+$\mathbf{T}(\mathbf{D}_f)$), as well as the weighting of local sparsity ($\mathbf{T}(\mathbb{S}_g)$+$\mathbf{T}(\mathbb{S}_f)$). Both attention/weighting settings contribute to the performance improvement, which validate the effectiveness of our MNEW design.

{\bf Performance varied by Distance and Sparsity.}
Since we target sparse point clouds segmentation towards the application of autonomous driving perception, it is interesting to know the performance for points with different distances (with respect to the LiDAR sensor) or sparsity (with respect to nearby points). Euclidean distances are computed in the geometry space, and sparsity is calculated using the normalized value of Equation (\ref{eq4}). Investigated on SemanticKITTI, we demonstrate distribution histograms of distance and sparsity, as well as result variations in Figure \ref{fig4}. We observe in Figure \ref{fig4a}, the performance occurs an abrupt elevation at $\approx$50 meters distance. We explicit that points in far away areas (e.g., $>$50m) are limited in quantity (refer to the distance distribution), but generally involved in well-leaned categories (e.g., road, building, terrain) which contain relatively large percentage of samples. In Figure \ref{fig4b}, as the sparsity increases, the performance starts to decrease until 0.7$\sim$0.8 and then increase. This is corresponded with the sparsity distribution, implying that the amount of samples affect the performance. Since sparsity distributions for dense datasets are relatively uniform, we infer it an essential reason for the effectiveness disparity of existing methods. It is also observed from Figure \ref{fig4} that MNEW achieves winning performance against other methods across the distance and sparsity distribution.


\section{Conclusion} \label{Conclusion}

In this work, we propose MNEW for sparse point clouds segmentation. MNEW inherits a dilation architecture to capture pointwise and global features, and involves multi-scale local semantics adopted from the hierarchical encoder-decoder structure. Neighborhood information is embedded in both static geometry and dynamic feature domain. The geometry distance, feature similarity, and local sparsity are computed and transformed as adaptive weighting factors. We obtain outstanding performances on both sparse point clouds. We believe this study will contribute to the application of LiDAR-based autonomous driving perception. 

In the continued work, one direction is to extend the model for joint semantic and instance segmentation, i.e., panoptic segmentation \cite{kirillov2019panoptic} for point clouds. The second ongoing direction is domain adaption to resolve the issue of cross-sensor disparity. The individual model could also be light-weighted and exported version as a LiDAR feature extractor, which is useful to fuse with other sensors such as radar and camera \cite{zheng2019gfd}.


{\small
	\bibliographystyle{ieee_fullname}
	\bibliography{egbib}
}

\end{document}